\documentclass{article}
\usepackage{spconf,amsmath,graphicx}
\usepackage[fontsize=9pt]{scrextend}

\usepackage{booktabs}
\usepackage{balance}

\usepackage{amsmath,epsfig,amssymb,amsthm,url,amsfonts}
\usepackage{algorithm}
\usepackage{algpseudocode}
\usepackage{multirow}
\usepackage{mdframed}
\usepackage{url}
\usepackage{enumitem}
\usepackage[makeroom]{cancel}
\usepackage{dblfloatfix}
\usepackage{array}
\usepackage{comment}
\usepackage{epstopdf}
\usepackage{float}
\usepackage{subfig}
\usepackage{breqn}
\usepackage{cite}
\usepackage{xcolor}
\usepackage{cases}
\usepackage{tabularx}
\usepackage[printonlyused,withpage]{acronym}
\usepackage{balance}
\usepackage{graphbox} 

\usepackage{hyperref}
\setlist[itemize]{label=$\triangleright$}
\newtheoremstyle{break}
{}
{}
{\itshape}
{}
{\bfseries}
{.}
{\newline}
{}
\theoremstyle{break}

\theoremstyle{definition}

\newcommand{\vect}[1]{\mathbf{#1}}
\newcommand{\bs}[1]{\boldsymbol{#1}}
\newcommand{\E}{\mathbb{E}}
\usepackage{url}

\makeatletter
\def\thmhead@plain#1#2#3{%
	\thmname{#1}\thmnumber{\@ifnotempty{#1}{ }\@upn{#2}}%
	\thmnote{ {\the\thm@notefont#3}}}
\let\thmhead\thmhead@plain
\makeatother

\graphicspath{{./figs/}}

\newcommand{\argmax}{\operatornamewithlimits{argmax}}

\newcommand{\lk}{ \left\{ }
\newcommand{\rk}{ \right\} }

\newcommand{\Hb}{{\bf H}}

\newcommand{\Wb}{{\bf W}}

\newcommand{\diag}{\mbox{{diag}}}

\newcommand{\Ib}{{\bf I}}

\newsavebox\mybox

\acrodef{SE}{speech enhancement}
\acrodef{STFT}{short-time Fourier transform}
\acrodef{STOI}{short-time objective intelligibility}
\acrodef{PSD}{power spectral density}
\acrodef{NMF}{non-negative matrix factorization}
\acrodef{AV}{audio-visual}
\acrodef{DNN}{deep neural network}
\acrodefplural{DNNs}{deep neural networks}
\acrodef{VAE}{variational auto-encoder}
\acrodefplural{VAEs}{variational auto-encoders}
\acrodef{CVAE}{conditional variational auto-encoder}
\acrodefplural{CVAEs}{conditional variational auto-encoders}
\acrodef{A-VAE}{audio VAE}
\acrodef{V-VAE}{visual VAE}
\acrodef{AV-CVAE}{audio-visual CVAE}
\acrodef{ROI}{region of interest}
\acrodef{MCMC}{Markov Chain Monte Carlo}
\acrodef{EM}{expectation-maximization}
\acrodef{MCEM}{Monte Carlo expectation-maximization}
\acrodef{TF}{time frequency}
\acrodef{ELBO}{evidence lower bound}
\acrodef{ROI}{region of interest}
\acrodef{LR}{Living Room}
\acrodef{SDR}{signal-to-distortion ratio}
\acrodef{PESQ}{perceptual evaluation of speech quality}
\acrodef{ASE}{audio speech enhancement}
\acrodef{VSE}{visual speech enhancement}
\acrodef{AVSE}{audio-visual speech enhancement}
\acrodef{SNR}{signal-to-noise ratio}
\acrodefplural{SNRs}{signal-to-noise ratios}
\acrodef{LSTM}{long short-term memory}
\acrodef{DNNs}{deep neural networks}
\acrodef{RNN}{recurrent neural network}

\title{Posterior Sampling Algorithms for Unsupervised Speech Enhancement with Recurrent Variational Autoencoder}
\name{%
Mostafa Sadeghi,
Romain Serizel %
\thanks{This work was supported by the French National Research Agency (ANR) under the project REAVISE (ANR-22-CE23-0026-01). Experiments presented in this paper were carried out using the Grid'5000 testbed, supported by a scientific interest group hosted by Inria, and including CNRS, RENATER, and several universities as well as other organizations (see https://www.grid5000.fr). }
}
\address{%
Université de Lorraine, CNRS, Inria, LORIA, F-54000 Nancy, France}
\begin{document}
%
\maketitle
\begin{abstract}
In this paper, we address the unsupervised speech enhancement problem based on recurrent variational autoencoder (RVAE). This approach offers promising generalization performance over the supervised counterpart. Nevertheless, the involved iterative variational expectation-maximization (VEM) process at test time, which relies on a variational inference method, results in high computational complexity. To tackle this issue, we present efficient sampling techniques based on Langevin dynamics and Metropolis-Hasting algorithms, adapted to the EM-based speech enhancement with RVAE. By directly sampling from the intractable posterior distribution within the EM process, we circumvent the intricacies of variational inference. We conduct a series of experiments, comparing the proposed methods with VEM and a state-of-the-art supervised speech enhancement approach based on diffusion models. The results reveal that our sampling-based algorithms significantly outperform VEM, not only in terms of computational efficiency but also in overall performance. Furthermore, when compared to the supervised baseline, our methods showcase robust generalization performance in mismatched test conditions.
\end{abstract}
\section{Introduction}
\label{sec:intro}
Speech enhancement is a fundamental signal processing technique, aiming to improve the quality and intelligibility of a noisy speech signal corrupted by acoustic noise \cite{vincent2018audio}. Over the past few years, and with the unprecedented success of deep learning, speech enhancement approaches have shifted from traditional statistical methods to data-driven approaches based on \ac{DNNs} \cite{wang2018supervised, xu2014regression,richter2023speech,yen2023cold,lu2022conditional}. Predominantly, current DNN-based speech enhancement techniques adopt a supervised (discriminative) paradigm, wherein a DNN is trained to map noisy speech inputs to their corresponding clean counterparts, leading to state-of-the-art performance. However, a notable challenge pervasive in these methods concerns generalization to test conditions not encountered during training, such as distinct noise types and noise levels that deviate from training conditions. 

In contrast, unsupervised speech enhancement methods based on deep generative models do not learn noise characteristics during the training process \cite{bando2018statistical, bie2022unsupervised, carbajal2021guided, bando2020adaptive}. Specifically, a deep generative model, most commonly based on the \ac{VAE} \cite{KingW14}, is trained solely on clean speech signals. This trained model then serves as a prior distribution for estimating clean speech from noisy input using an \ac{EM} approach. This gives them a generalization advantage over discriminative approaches. However, unsupervised methods remain significantly less explored than their supervised counterparts and suffer from some challenges, including their notably high computational complexity. This complexity originates from the iterative \ac{EM} process during inference, which requires sampling from an intractable posterior distribution. For instance, the current state-of-the-art method for unsupervised speech enhancement relies on recurrent \ac{VAE} (RVAE) \cite{leglaive2020recurrent,bie2022unsupervised}, as a dynamical and more efficient version of the standard VAE. This approach adopts a variational \ac{EM} (VEM) strategy, involving the fine-tuning of the trained encoder at each EM iteration on the input noisy speech. Its computational complexity thus grows with the complexity (number of parameters) of the encoder.

To address this issue, we propose alternative, more efficient posterior sampling-based methods for speech enhancement with RVAE. The first approach extends the Langevin dynamics \ac{EM} (LDEM) method for standard, non-dynamical \ac{VAE} presented in \cite{sadeghi2023fast} to RVAE. This technique involves sampling from the intractable posterior using gradient descent steps combined with Gaussian noise injection. Additionally, we develop a Metropolis-Hastings (MH) sampling technique \cite{bishop2006pattern}, relying on a proposal and acceptance/rejection mechanism, to generate a sequence of samples. Lastly, a Metropolis-adjusted Langevin algorithm (MALA) \cite{roberts2002langevin} is proposed, combining the strengths of both LDEM and MH methods. We assess the effectiveness of these algorithms for RVAE-based speech enhancement by comparing them to the VEM method and a state-of-the-art supervised speech enhancement approach based on diffusion models \cite{richter2023speech}, in both matched and mismatched test conditions. The results demonstrate that our proposed speech enhancement algorithms outperform VEM significantly in terms of performance and computational efficiency. Furthermore, they exhibit more robust generalization performance when compared to the supervised baseline method.

The paper is organized as follows: In Section~\ref{sec:vae}, we present an overview of unsupervised speech enhancement based on RVAE. Section~\ref{sec:prop} introduces the proposed posterior sampling methods. Our experimental results are detailed in Section~\ref{sec:exp}. Lastly, Section~\ref{sec:conc} provides some conclusions.
\vspace{-2mm}
\section{Background}
\label{sec:vae}
\subsection{RVAE as a deep speech prior}
We denote by $\vect{s}\triangleq \vect{s}_{1:T}=\{\vect{s}_1,\ldots,\vect{s}_T\}$ a sequence of clean speech time-frequency representations computed using \ac{STFT}, where ${\vect{s}_t = [s_{ft}]_{f=1}^F \in \mathbb{C}^F}$. RVAE \cite{leglaive2020recurrent}, as a latent variable-based deep generative model, considers the following generative model for the speech time frames $\vect{s}$:
\begin{equation}\label{eq:rvae_gen}
p_\theta(\vect{s},\vect{z})=\prod_{t=1}^T p_\theta(\vect{s}_t|\vect{z})p(\vect{z}_t)
\end{equation}
where $\vect{z}=\{\vect{z}_1,\ldots,\vect{z}_T\}$, $\vect{z}_t\in\mathbb{R}^L$ ($L\ll F$), are low-dimensional latent variables associated with $\vect{s}$. Moreover,
\begin{equation}\label{eq:rvae_dec}
    p_\theta(\vect{s}_t|\vect{z})=\mathcal{N}_c(\boldsymbol{0}, \diag(\bs{v}_{\theta,t}(\vect{z})))
\end{equation}
is a circularly-symmetric complex Gaussian distribution, with a diagonal covariance matrix whose entries, given by $\bs{v}_{\theta,t}(\vect{z})$, are modeled by a \ac{RNN}, called decoder. Here, $\bs{v}_{\theta,t}(\vect{z})$ refers to the output at time
frame $t$ of the RNN with $\vect{z}$ as the input. This dynamical modeling makes RVAE more efficient than the standard \ac{VAE}. Similar to \ac{VAE}, the prior $p(\vect{z}_t)$ is set to a standard Gaussian distribution.

Training the generative model \eqref{eq:rvae_gen} involves learning the RNN parameters $\theta$ following an \ac{EM} procedure. The intractable posterior $p_\theta(\vect{z}|\vect{s})$ is approximated with a parametric Gaussian distribution as follows
\begin{equation}
    q_\phi(\vect{z}|\vect{s})=\prod_{t=1}^T q_\phi(\vect{z}_t|\vect{z}_{1:t-1}, \vect{s}_{t:T}),
\end{equation}
where similarly as in \eqref{eq:rvae_dec}, the mean and variance are modeled via an RNN, called encoder, with parameters denoted $\phi$. The encoder and decoder parameters, i.e., $\{\theta, \phi\}$, are then jointly learned by optimizing an evidence lower-bound \cite{KingW14}.
\subsection{Variational EM for speech enhancement}
The observation model for speech enhancement is assumed to be $\vect{x}_t = \vect{s}_t+\vect{b}_t$, with $\vect{b}_t$ corresponding to noise. As a statistical model for clean speech $\vect{s}_t$, the pretrained RVAE model, i.e., $p_\theta(\vect{s}, \vect{z})$ is used. Moreover, noise is modeled based on a \ac{NMF} model \cite{bando2018statistical}, where a circularly symmetric Gaussian form $p_\psi(\vect{b}_t)\sim \mathcal{N}_c(\bs{0}, \diag([\Wb\Hb]_t))$ is considered. The non-negative matrices $\Wb,\Hb$ form the noise parameters $\psi$ to be learned from $\vect{x}$. This is done following an \ac{EM} approach, that is
\begin{equation}\label{eq:inference}
\psi^*=\argmax_{\psi}~\E_{p_\psi(\vect{z}|\vect{x})}\lk\log p_\psi(\vect{x}, \vect{z})\rk,
\end{equation}
where $p_\psi(\vect{x}, \vect{z})=\prod_{t} p_\psi(\vect{x}_t| \vect{z})p(\vect{z}_t)$, with likelihood computed as $p_\psi(\vect{x}_t| \vect{z})=\mathcal{N}_c(\boldsymbol{0}, \diag(\bs{v}_{\theta,t}(\vect{z})+[\Wb\Hb]_t))$ \cite{leglaive2020recurrent}. Here, the posterior $p_\psi(\vect{z}|\vect{x})$ in \eqref{eq:inference}, needed for the E-step, is intractable to compute. The variational \ac{EM} (VEM) approach proposed in \cite{leglaive2020recurrent} fine-tunes the pretrained encoder $q_\phi(\vect{z}|\vect{s})$ on $\vect{x}$ at each E-step, to serve as an approximation of $p_\psi(\vect{z}|\vect{x})$. This approach aligns with the principles of standard variational inference methods. Then, at the M-step, using latent variables sampled from the approximate posterior, the NMF parameters are updated by optimizing \eqref{eq:inference}. Once the EM steps converge, the speech signal is estimated as $\hat{\vect{s}} = \E_{p_{\psi^*}(\vect{s}|\vect{x})} \lk \vect{s}\rk$.
\begin{algorithm}[t!]
\caption{EM-based speech enhancement}
\label{alg:em-se}
\begin{algorithmic}[1]
\State \textbf{Inputs:} $\vect{x}=\lk \vect{x}_t\rk_{t=1}^{{T}}$ ({\footnotesize noisy STFT data}), $\mathcal{H}$ ({\footnotesize hyperparameters}).
\State \textbf{Initialize:} $\vect{z}=\lk \vect{z}_t\rk_{t=1}^{{T}}$, $\psi=$\{$\Wb$, $\Hb$\}.
\For{$j=1,\cdots,J$}
\State \textbf{E-step:}\hfil \hspace{-4.5mm} $ \vect{z} \leftarrow \texttt{Sampler}_{\psi}(\vect{z}, \mathcal{H})$
\State \textbf{M-step:}\hfil $\psi\leftarrow\argmax_{\psi}\log p_\psi(\vect{x}| \vect{z})$

\EndFor

\State \textbf{Clean speech estimation:} 
$\hat{\vect{s}}=\lk\frac{\bs{v}_{\theta,t}(\vect{z})}{\bs{v}_{\theta,t}(\vect{z}) + [\Wb\Hb]_t}\odot \vect{x}_t \rk_{t=1}^{{T}}$
\end{algorithmic}
\end{algorithm}
\section{Posterior sampling algorithms}
\label{sec:prop}
In this section, we present our EM-based speech enhancement frameworks, utilizing RVAE as a deep speech prior. These frameworks share a common structure but vary in the E-step, where each employs a distinct strategy to draw samples from the intractable posterior $p_\psi(\vect{z}|\vect{x})$. We provide a concise summary of the overall speech enhancement process in Algorithm~\ref{alg:em-se}. Specifically, the first approach extends the LDEM method, as proposed in \cite{sadeghi2023fast}, to RVAE. The second approach utilizes the Metropolis-Hastings sampling algorithm, while the third algorithm is a Metropolis-adjusted version of LDEM.

\subsection{Langevin dynamics (LD)}
In the conventional VAE-based speech enhancement method described in \cite{sadeghi2023fast}, the process of sampling from the posterior distribution is carried out independently for each latent variable. To capture temporal dependencies, a total variation (TV) regularization term is introduced. However, in the context of RVAE, latent variables are naturally interconnected through an RNN model, making the TV regularization term redundant. 

Langevin dynamics enables the generation of a sequence of samples from the posterior distribution $p_\psi(\vect{z}|\vect{x})$ solely using its score function, defined as follows:
\begin{align}
    f_{\psi}(\vect{z})&=\nabla_{\vect{z}} \log p_\psi(\vect{z}|\vect{x}) \nonumber\\ &= \nabla_{\vect{z}} \Big(\log p_\psi(\vect{x}|\vect{z}) + \log p(\vect{z}) \Big)\nonumber\\
    &= \nabla_{\vect{z}} \Big(\sum_{t=1}^T \log p_\psi(\vect{x}_t|\vect{z}) + \log p(\vect{z}_t) \Big).
\end{align}
In contrast to VAE, this score function cannot be decomposed over individual latent variables, meaning that $f_{\psi}(\vect{z})\neq \sum_{t} f_{\psi}(\vect{z}t)$. Consequently, each $\vect{z}_t$ must be sampled individually, akin to the sequential Gibbs sampling procedure \cite{bishop2006pattern}. This sequential approach would significantly increase complexity. Instead, we adopt a parallel sampling strategy, wherein all latent variables are sampled simultaneously. Furthermore, following the methodology employed in LDEM for VAE, we generate multiple samples for each latent variable to obtain a more robust and efficient approximation of the expectation in \eqref{eq:inference}. Therefore, starting from $\vect{z}=\left(\vect{z}_{1}, \cdots, \vect{z}_{{T}}\right)$, we initially draw $M$ distinct samples (states) for each latent variable $\vect{z}_{t}$, denoted as $\Bar{\vect{z}}^{(0)}= \lk\vect{z}_{t,i}^{(0)}\rk_{t,i} $, with $t=1,\ldots T$ and $i=1,\ldots M$, using a random walk approach by sampling from the following proposal distribution:
\begin{equation}
\label{eq:proposal}
    \vect{z}_{t,i}^{(0)}|\vect{z}_{t} \sim \mathcal{N}(\vect{z}_{t}, \sigma^2\Ib),~~~\forall t, i
\end{equation}
or $\vect{z}_{t,i}^{(0)} = \vect{z}_t + \sigma \bs{\epsilon}_{t,i}$, where $\bs{\epsilon}_{t,i}\sim\mathcal{N}(\bs{0}, \Ib)$ and $\sigma^2>0$. The next states are produced by sampling from the following distribution:
\begin{equation}\label{eq:prop_ld}
    \vect{z}_{t,i}^{(k)}|\Bar{\vect{z}}^{(k-1)} \sim \mathcal{N}(\vect{z}_{t,i}^{(k-1)} + \frac{\eta}{2} f_\psi(\Bar{\vect{z}}^{(k-1)}), \eta \Ib),
\end{equation}
or, equivalently
\begin{equation}\label{eq:ld}
    \vect{z}_{t,i}^{(k)} = \vect{z}_{t,i}^{(k-1)} + \frac{\eta}{2} f_\psi(\Bar{\vect{z}}^{(k-1)}) + \sqrt{\eta} \bs{\zeta}_{t,i},
\end{equation}
where $\bs{\zeta}_{t,i}\sim\mathcal{N}(\bs{0}, \Ib)$, and $\eta>0$ is a step size. The added noise introduces stochasticity, which enhances exploration within the high-density regions of the posterior. The generated samples converge to the true posterior distribution under some regularity conditions \cite{welling2011bayesian}. The LD sampler is summarized in Algorithm~\ref{alg:ld}.

\begin{algorithm}[t!]
\caption{LD sampler}
\label{alg:ld}
\begin{algorithmic}[1]
\State \textbf{Inputs:} $\Bar{\vect{z}}^{(0)} = \lk \vect{z}_{t,i}^{(0)}\rk_{t,i}$, $\mathcal{H}$ ({\footnotesize hyperparameters}).
\For{$k=1,\cdots,K$}
\State $\bs{\zeta} = \lk\bs{\zeta}_{t,i} \rk_{t,i}$, with $\bs{\zeta}_{t,i}\sim\mathcal{N}(\bs{0}, \Ib)$
\State $    \Bar{\vect{z}}^{(k)} = \Bar{\vect{z}}^{(k-1)} + \frac{\eta}{2} f_\psi(\Bar{\vect{z}}^{(k-1)}) + \sqrt{\eta} \bs{\zeta}$,
\EndFor

\State \textbf{Output:} $\Bar{\vect{z}}^{(K)} = \lk \vect{z}_{t,i}^{(K)}\rk_{t,i}$
\end{algorithmic}
\end{algorithm}

\begin{algorithm}[t!]
\caption{MH sampler}
\label{alg:mh}
\begin{algorithmic}[1]
\State \textbf{Require:} ${\vect{z}}^{(0)} = \lk \vect{z}_{t}^{(0)}\rk_{t}$, $\mathcal{H}$ ({\footnotesize hyperparameters}).
\For{$k=1,\cdots,K$}
\State $\bs{\zeta} = \lk\bs{\zeta}_{t} \rk_{t}$, with $\bs{\zeta}_{t}\sim\mathcal{N}(\bs{0}, \Ib)$
\State $    \Tilde{\vect{z}}^{(k)} = {\vect{z}}^{(k-1)} + \sqrt{\eta} \bs{\zeta}$,
\State Accept $\Tilde{\vect{z}}^{(k)}_t$ ($\forall t$) according to \eqref{eq:alpha_mh}
\EndFor

\State \textbf{Output:} $\Bar{\vect{z}} = \lk \vect{z}^{(k)}\rk_{k> k_{\texttt{{\tiny burn-in}}}}$
\end{algorithmic}
\end{algorithm}

\begin{algorithm}[t!]
\caption{MALA sampler}
\label{alg:mala}
\begin{algorithmic}[1]
\State \textbf{Require:} ${\vect{z}}^{(0)} = \lk \vect{z}_{t}^{(0)}\rk_{t}$, $\mathcal{H}$ ({\footnotesize hyperparameters}).
\For{$k=1,\cdots,K$}
\State $\bs{\zeta} = \lk\bs{\zeta}_{t} \rk_{t}$, with $\bs{\zeta}_{t}\sim\mathcal{N}(\bs{0}, \Ib)$
\State $    \Tilde{\vect{z}}^{(k)} = {\vect{z}}^{(k-1)} + \frac{\eta}{2} f_\psi({\vect{z}}^{(k-1)}) + \sqrt{\eta} \bs{\zeta}$,
\State Accept $\Tilde{\vect{z}}^{(k)}_t$ ($\forall t$) according to \eqref{eq:alpha_mala}
\EndFor

\State \textbf{Output:} $\Bar{\vect{z}} = \lk \vect{z}^{(k)}\rk_{k> k_{\texttt{{\tiny burn-in}}}}$
\end{algorithmic}
\end{algorithm}

\subsection{Metropolis-Hastings (MH)}
 Metropolis-Hastings (MH) \cite{robert1999monte} is a Markov chain Monte Carlo (MCMC) sampling technique for generating a sequence of samples from a probability distribution. It begins with initial states and iteratively proposes new states using a typically Gaussian proposal distribution. Candidate states are accepted or rejected based on defined probabilities.

For RVAE, similarly as done in the LD sampler, the MH algorithm generates samples for all the latent variables in parallel. More precisely, starting from some initial states $\vect{z}^{(0)}=\lk \vect{z}_{1}^{(0)}, \cdots, \vect{z}_{{T}}^{(0)}\rk$, at the $k$-th iteration, a sequence of candidate states, denoted $\Tilde{\vect{z}}^{(k)}$, are sampled from the following proposal distribution
\begin{equation}\label{eq:proposal_mh}
    \Tilde{\vect{z}}^{(k)}_{t} | \vect{z}^{(k-1)}_{t}\sim \mathcal{N}(\vect{z}^{(k-1)}_{t}, \sigma^2\Ib),~~~\forall t.
\end{equation}
Each candidate state $\Tilde{\vect{z}}^{(k)}_{t}$ in the sequence $\Tilde{\vect{z}}^{(k)}$ is then accepted according to the following probability:
\begin{equation}\label{eq:alpha_mh}
\alpha_t = \min\Big(1, \frac{p_\psi(\vect{x}_t|\Tilde{\vect{z}}^{(k)}) p(\Tilde{\vect{z}}^{(k)}_{t})}{p_\psi(\vect{x}_t|\vect{z}^{(k-1)}) p(\vect{z}^{(k-1)}_{t})} \Big)
\end{equation}
Let $u_t$ be drawn from the continuous uniform distribution over $[0,1]$. Then, if $u_t \leq \alpha_t$, the proposal is accepted and ${\vect{z}}^{(k)}_{t}\leftarrow\Tilde{\vect{z}}^{(k)}_{t}$. Otherwise, the current state is retained~${\vect{z}}^{(k)}_{t}\leftarrow{\vect{z}}^{(k-1)}_{t}$. A key observation here is that the acceptance ratios, $\alpha_1,\ldots,\alpha_T$, are computed in parallel, with the same current likelihood  $p_\phi(\vect{x}|\vect{z}^{(k-1)})$ for all the samples. Once a sufficient number of iterations is performed, the initial samples corresponding to the so-called burn-in period are discarded. The overall MH sampler is summarized in Algorithm~\ref{alg:mh}. 
\subsection{Metropolis-Adjusted Langevin Algorithm (MALA)}

The Metropolis-Adjusted Langevin Algorithm (MALA) \cite{roberts2002langevin} aims at combining the MH and LD mechanisms to achieve a more efficient exploration of the target distribution. MALA follows the same steps as MH with the difference that the candidate states are generated using a proposal distribution guided by LD. More precisely, the proposal distribution takes a similar form as \eqref{eq:prop_ld}, except for the fact that here we do not generate multiple samples per latent variable:
\begin{equation}
        \Tilde{\vect{z}}_t^{(k)}|\vect{z}_t^{(k-1)} \sim \mathcal{N}(\vect{z}_t^{(k-1)} + \frac{\eta}{2} f_\psi(\vect{z}_t^{(k-1)}), \eta \Ib),
\end{equation}
Nevertheless, in contrast to LD sampler, which always updates the states according to the update rule \eqref{eq:prop_ld}, MALA considers the updated states as candidates, similar to MH, and accepts them according to the following probability
\begin{equation}\label{eq:alpha_mala}
\alpha_t = \min\Big(1, \frac{p_\psi(\vect{x}_t|\Tilde{\vect{z}}^{(k)}) p(\Tilde{\vect{z}}^{(k)}_{t}) q({\vect{z}}^{(k)}|\Tilde{\vect{z}}^{(k)})}{p_\psi(\vect{x}_t|\vect{z}^{(k-1)}) p(\vect{z}^{(k-1)}_{t})q(\Tilde{\vect{z}}^{(k)}|{\vect{z}}^{(k)})} \Big)
\end{equation}
where
\begin{equation}
    q(\vect{u}|\vect{v}) \propto \exp\Big( -\frac{1}{2\eta} \|\vect{u}-\vect{v}-\frac{\eta}{2} f(\vect{v}) \|^2\Big)
\end{equation}
is the transition probability density from $\vect{v}$ to $\vect{u}$. Unlike the basic MH approach, MALA often suggests moves towards regions of higher probability, thus increasing the probability of their acceptance. The overall MALA sampler is described in Algorithm~\ref{alg:mala}. 

\section{Experiments}
\label{sec:exp}
\subsection{Baselines}
This section presents and discusses the speech enhancement results of the proposed EM-based approaches, i.e., LDEM, MHEM, and MALAEM, with RVAE \cite{leglaive2020recurrent} as the generative model. We compare against the VEM method\footnote{\url{https://github.com/XiaoyuBIE1994/DVAE_SE/}} \cite{leglaive2020recurrent,bie2022unsupervised}, and SGMSE+\footnote{\url{https://github.com/sp-uhh/sgmse}} \cite{richter2023speech}, as a state-of-the-art diffusion-based speech enhancement method. 

\subsection{Evaluation metrics}
To evaluate the speech enhancement performance, we use standard metrics, including the extended short-time objective intelligibility (ESTOI) measure~\cite{jensen2016algorithm}, ranging in $[0,1]$, the perceptual evaluation of speech quality (PESQ) metric~\cite{Rix2001pesq}, ranging in $[-0.5,4.5]$, and the scale-invariant signal-to-distortion ratio (SI-SDR) metric \cite{le2019sdr} in dB. For all these metrics, the higher the better. Moreover, as a rough measure of the computational complexity of different methods, we report the  average real-time factor (RTF), which is defined as the time (in seconds) required to enhance one second of speech signal. Our experiments were conducted using a machine with an AMD EPYC 7351 CPU and an NVIDIA Tesla T4 GPU.

\subsection{Model architectures}
We utilized pretrained models from the respective code repositories for both RVAE and SGMSE+. In the RVAE architecture, the encoder and decoder employ bidirectional long short-term memory (BLSTM) networks with an internal state dimension of 128, and the latent space is of dimension $L=16$. The input data consists of STFT power spectrograms with a dimension of $F=513$. This model was trained on the training subset of the Wall Street Journal (WSJ0) corpus. The architecture of SGMSE+ is detailed in \cite{richter2023speech}, and its core network is based on the Noise Conditional Score Network (NCSN++) \cite{song2020score}, adapted for processing complex-valued STFT features. The overall model was trained using the same clean training utterances as RVAE, combined with noise signals from the CHiME3 dataset \cite{barker2015third}. The input data for SGMSE+ comprises STFT representations with $F=256$, and training was conducted on sequences with a length of $T=256$, as opposed to $T=50$ used for RVAE.

\subsection{Parameter settings}
For the inference parameters of SGMSE+, we adhered to their default values. In the case of RVAE-based methods, we conducted a total of $J=100$ EM iterations. The learning rate for VEM, with the Adam optimizer, as well as for LDEM and MALAEM (denoted as $\eta$), was consistently set at $0.005$. For LDEM and VEM, we empirically selected $K=1$, while for MHEM and MALAEM, we opted for $K=10$, and included a burn-in period of $k_{\texttt{{\tiny burn-in}}}=5$. Additionally, we set $\sigma^2=0.02$ in both \eqref{eq:proposal} and \eqref{eq:proposal_mh}.

\subsection{Evaluation datasets}
For performance evaluation, we used the test set from of the WSJ0-QUT corpus \cite{leglaive2020recurrent}, created by mixing clean speech signals from WSJ0 (distinct speakers from training) with noise signals from the QUT-NOISE corpus \cite{dean2015qut}. It includes \textit{Café}, \textit{{Home}}, \textit{Street}, and \textit{Car} noise types, with signal-to-noise ratio (SNR) values of $-5$~dB, $0$~dB, and $5$~dB. The test set amounts to 651 noisy speech signals with a total duration of 1.5 hours. Furthermore, we evaluated the generalization performance of various methods under mismatched conditions by incorporating pre-computed noisy versions of the TCD-TIMIT data as introduced in \cite{abdelaziz2017ntcd}.  The noisy test set that we used includes \textit{\ac{LR}}  (from the second CHiME challenge \cite{vincent2013second}), \textit{White}, \textit{Car}, and \textit{Babble} (from the RSG-10 corpus \cite{steeneken1988description}), with SNR\footnote{Here, the protocol used to compute SNR is different than the one used in \cite{leglaive2020recurrent}.} values of $-5$~dB, $0$~dB, and $5$~dB, yielding 540 test speech signals, with a total duration of about 0.75 hours. 

\subsection{Results}
The average speech enhancement metrics, under both matched (WSJ0-QUT) and mismatched (TCD-TIMIT) conditions, are presented in Table~\ref{tab:se_res}, with the associated average RTF values reported in Table~\ref{tab:runtime}. From these results, we can make several observations. First of all, among the RVAE-based algorithms, the proposed posterior sampling methods outperform VEM with a significant margin. The performance gap is even more remarkable in the mismatched conditions, which demonstrates that VEM is not as generalizable as our proposed methods. This could be due to the fact that VEM relies on fine-tuning the trained encoder on the new data, which might not be efficient. For instance, in the mismatched condition, MALAEM outperforms VEM by around \textbf{3 dB} in SI-SDR, \textbf{0.19} in PESQ, and \textbf{0.07} in ESTOI. Furthermore, the LDEM algorithm consistently stands out with the highest or second-highest scores in all three metrics under both test conditions. The observation that LDEM outperforms MALAEM could be because of their different sampling strategies. That is, LDEM creates multiple parallel sequences of samples at each E-step, whereas MALAEM has only one sequential sequence of samples, where the final retained samples are chosen based on a probabilistic acceptance strategy.

On the other hand, SGMSE+, as the supervised baseline, showcases remarkable performance in the matched condition, achieving much higher speech enhancement metrics than those of the unsupervised RVAE-based methods. Nevertheless, when tested in the mismatched condition, the performance of SGMSE+ drops significantly, under-performing our proposed methods with a large margin. This confirms the generalization dilemma of supervised methods.

In terms of computational complexity at inference time, all the proposed methods achieve much smaller RTF values than VEM, making them more applicable. In particular, the LDEM algorithm demonstrates a competitive RTF of \textbf{0.21 sec}, as compared to \textbf{12.55 sec} for VEM and \textbf{3.85 sec} for SGMSE+, showcasing its high computational efficiency in enhancing speech signals

\begin{table}[t!]
\centering
	\caption{Speech enhancement results under matched (WSJ0-QUT) and mismatched (TCD-TIMIT) test conditions.}
\resizebox{0.485\textwidth}{!}{
\begin{tabular}{|l|l|c|c|c|}
\hline
 \multicolumn{2}{|l|}{Metric} & {SI-SDR (dB)} & {PESQ} & {ESTOI} \\
\hline
\multicolumn{2}{|l|}{Input (\textbf{WSJ0-QUT})} & -2.60~$\pm$~0.16   & 1.83~$\pm$~0.02 & 0.50~$\pm$~0.01 \\ \hline
\multirow{4}{*}{RVAE} & VEM \cite{bie2022unsupervised} & 4.5~$\pm$~0.21 & {2.21}~$\pm$~0.02 & 0.60~$\pm$~0.01 \\ 
 & MHEM & 5.15~$\pm$~0.20  & {2.24}~$\pm$~0.02 & 0.62~$\pm$~0.01 \\ 
  & MALAEM & 5.52~$\pm$~0.21  & {2.28}~$\pm$~0.02 & 0.62~$\pm$~0.01  \\ 
 & LDEM & \underline{5.58}~$\pm$~0.20  & \underline{2.32}~$\pm$~0.02 & \underline{0.63}~$\pm$0.01\\ \hline
 \multicolumn{2}{|l|}{SGMSE+ \cite{richter2023speech}} & \textbf{9.41}~$\pm$~0.18   & \textbf{2.66}~$\pm$~0.02 & \textbf{0.77}~$\pm$~0.01   \\ \hline\hline
 \multicolumn{2}{|l|}{Input (\textbf{TCD-TIMIT})}& -8.74~$\pm$~0.29 & 1.84~$\pm$~0.02 & 0.35~$\pm$~0.01  \\ \hline
\multirow{4}{*}{RVAE} & VEM \cite{bie2022unsupervised} & 1.44~$\pm$~0.30 & 2.02~$\pm$~0.02 & 0.35~$\pm$~0.01\\ 
    & MHEM & 3.72~$\pm$~0.27 & 2.12~$\pm$~0.02 & \textbf{0.42}~$\pm$~0.01 \\ 
    & MALAEM & \textbf{4.49}~$\pm$~0.29 & \textbf{2.21}~$\pm$~0.02 & \textbf{0.42}~$\pm$~0.01 \\ 
  & LDEM & \underline{4.18}~$\pm$~0.29 & \textbf{2.21}~$\pm$~0.02 & \textbf{0.42}~$\pm$~0.01\\ \hline
   \multicolumn{2}{|l|}{SGMSE+ \cite{richter2023speech}}& -3.97~$\pm$~0.41  & 2.04~$\pm$~0.02 & 0.38~$\pm$~0.01    \\ \hline
\end{tabular}}
\label{tab:se_res}
\end{table}

\begin{table}[t!]
\centering
	\caption{RTF values (average processing time per 1-sec speech).}
\resizebox{0.48\textwidth}{!}{
\begin{tabular}{|c|c|c|c|c|}
\hline
VEM \cite{bie2022unsupervised} & MHEM & MALAEM & LDEM & SGMSE+ \cite{richter2023speech}     \\ \hline
12.55~$\pm$~0.01 & \underline{0.92}~$\pm$~0.01 & 2.49~$\pm$~0.01 & \textbf{0.21}~$\pm$~0.01 & 3.85~$\pm$~0.01 \\
\hline
\end{tabular}}
\label{tab:runtime}
\end{table}

\section{Conclusions}\label{sec:conc}

In this paper, we present new posterior sampling techniques for EM-based unsupervised speech enhancement using RVAE. These methods serve as viable alternatives to the computationally intensive variational inference-based VEM approach. Our experimental results illustrate the efficiency of the proposed techniques—LDEM, MHEM, and MALAEM—which not only significantly reduce computational complexity but also outperform VEM. Notably, the LDEM algorithm demonstrates high efficiency and competitive enhancement outcomes. In contrast, the supervised baseline, SGMSE+, excels under matched conditions but faces challenges in mismatched scenarios, highlighting generalization limitation of supervised methods. In summary, our proposed methods offer a promising avenue for efficient and effective unsupervised speech enhancement.


\bibliographystyle{IEEEbib-abbrev}
\bibliography{mybib}

\end{document}